\documentclass{article}
\pdfoutput=1

\usepackage{PRIMEarxiv}

\usepackage[utf8]{inputenc} 
\usepackage[T1]{fontenc}    
\usepackage[bookmarks=false]{hyperref}      
\usepackage{url}            
\usepackage{booktabs}       
\usepackage{amsfonts}       
\usepackage{tabularx}
\usepackage{nicefrac}       
\usepackage{microtype}      
\usepackage{lipsum}		
\usepackage{graphicx}
\usepackage{multirow}
\usepackage[numbers]{natbib}
\usepackage{doi}
\hypersetup{
     colorlinks=true,
     linkcolor=blue,
     filecolor=magenta,      
     urlcolor=cyan,
     }

\title{PART: Pre-trained Authorship Representation Transformer}

\author{ \href{https://scholar.google.es/citations?user=5XOhXooAAAAJ}{\includegraphics[scale=0.02]{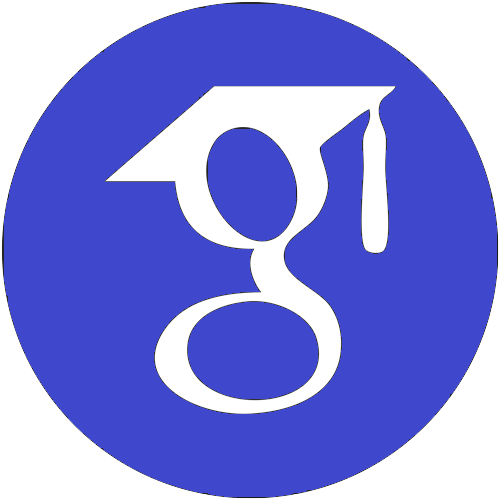}\hspace{1mm}Javier Huertas-Tato} \\
	Departamento de Sistemas Informáticos\\
	Universidad Politécnica de Madrid\\
	Spain, Madrid 28031 \\
	\texttt{javier.huertas.tato@upm.es} \\
	\And
	\href{https://scholar.google.com/citations?user=b3J9VRsAAAAJ}{\includegraphics[scale=0.02]{g_scholar.png}\hspace{1mm}Alejandro Martín} \\
	Departamento de Sistemas Informáticos\\
	Universidad Politécnica de Madrid\\
	Spain, Madrid 28031 \\
	\texttt{alejandro.martin@upm.es} \\
	\And
	\href{https://scholar.google.com/citations?user=fpf6EDAAAAAJ}{\includegraphics[scale=0.02]{g_scholar.png}\hspace{1mm}David Camacho} \\
	Departamento de Sistemas Informáticos\\
	Universidad Politécnica de Madrid\\
	Spain, Madrid 28031 \\
	\texttt{david.camacho@upm.es} \\
}

\hypersetup{
pdftitle={PART: Pre-trained Authorship Representation Transformer},
pdfsubject={cs.CL},
pdfauthor={Javier Huertas-Tato, Alvaro Huertas-Garcia, Alejandro Martin, David Camacho},
pdfkeywords={Transformers, Authorship profiling, ...},
}

\begin{document}
\maketitle

\begin{abstract}
Authors writing documents imprint identifying information within their texts: vocabulary, registry, punctuation, misspellings, or even emoji usage. Previous works use hand-crafted features or classification tasks to train their authorship models, leading to poor performance on out-of-domain authors. Using stylometric representations is more suitable, but this by itself is an open research challenge. In this paper, we propose PART, a contrastively trained model fit to learn \textbf{authorship embeddings} instead of semantics. We train our model on ~1.5M texts belonging to 1162 literature authors, 17287 blog posters and 135 corporate email accounts; a heterogeneous set with identifiable writing styles. We evaluate the model on current challenges, achieving competitive performance. We also evaluate our model on test splits of the datasets achieving zero-shot 72.39\% accuracy when bounded to 250 authors, a 54\% and 56\% higher than RoBERTa embeddings. We qualitatively assess the representations with different data visualizations on the available datasets, observing features such as gender, age, or occupation of the author.
\end{abstract}


\section{Introduction}
Authorship of textual pieces influences the contents of a work in varied ways. Each author has an education, has grown up in a socio-economic context, has lived through different experiences and has his/her own writing style, which we hypothesize influences whatever they have produced. Thus, proper methods of analysing authorship would allow for robust attribution of authors with wide range of applications. This need has been identified in the literature, where there are several applications for textual authorship~\cite{stamatatos2009survey, koppel2009computational}, with great potential in domains such as forensics~\cite{nini2015authorship}, mental health~\cite{Munoz2022Sep}, online social network analysis~\cite{bhargava2013stylometric,layton2010authorship} or disinformation spreading~\cite{potthast2018stylometric,bevendorff2020overview}.

While feature extraction is the usual approach to understand authorship or style~\cite{argamon2009automatically}, lack of flexibility to newer authors and limited amount of features among other weaknesses, may draw back the applicability of authorship attribution to practical domains. Analyzing parts-of-speech which are not included in the initial feature space might be difficult if not impossible altogether. We are concerned with the capacity of manual feature extraction for this task.

A more general-purpose approach to authorship would be to generate representations akin to document embeddings~\cite{gomez2018document}. Embeddings can extract meaning from textual speech by analyzing content and style, frequently ignoring contextual external information. These embeddings are a numerical representation of any given document that can be analysed with any appropriate method. Modern document embeddings can be extracted thanks to developments in Natural Language Processing (NLP), through the likes of pre-trained BERT~\cite{Devlin2018Oct}. Transformers have meant a breakthrough in text analysis and solved similar open-ended problems. In particular, for this authorship task, our interest is in encoders. For instance, BERT can be combined with text-based features to improve results~\cite{fabien2020bertaa}, generating tokens to classify which author would have written it~\cite{barlas2020cross} or mixing pooling techniques to merge the embeddings of several documents to profile the author~\cite{huertas2021profiling}.

Following these approaches, we aim to take authorship analysis a step further with a concept called \textbf{authorship embeddings}. We present a Pretrained Authorship Representation Transformer (PART)\footnote{Code repository: \href{https://github.com/jahuerta92/authorship-embeddings}{https://github.com/jahuerta92/authorship-embeddings}} with style characterization capabilities to include general features of speech from authors, allowing to represent a concrete author style with a numerical vector. Our PART embedding generation can be used to project words, sentences and documents into a contextually and style-aware hyper-space. We aim to achieve the same for authorship.

\begin{figure}[htbp]
\centerline{\includegraphics[width=0.75\linewidth]{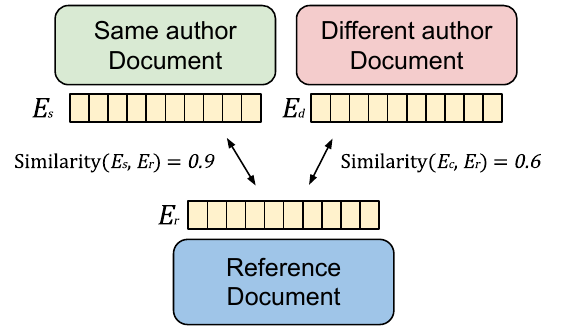}}
\caption{Example of a comparison of authorship embeddings. $E_r$, $E_s$, $E_d$ represent authorship embeddings from a reference document, a document whose author is the same as the reference and a text from a completely different author. When compared with a similarity function, the related document similarity should be higher than an unrelated text. }
\label{fig:summary}
\end{figure}

A suitable method to build representations is contrastive learning. It has been used in other information modalities such as image or audio to great success and recently, it has seen much use with CLIP~\cite{radford2021learning}. This work is able to align the embeddings of images with their annotated text, which is a similar task to our own. By contrasting a document embedding with another document embedding of the same author, the remaining representation will be aligned towards the author and not the contents of the document. A conceptual representation of our approach can be found in Fig.~\ref{fig:summary}. 

We present the following contributions in this article:
\begin{enumerate}
  \item This paper outlines a new contrastive learning approach to address authorship attribution.

  \item This article presents a stylometric sentence embedding generator. When extended to documents via average pooling, projections remain consistent, smoothing the length bottleneck.
  
  \item We compute author embeddings with zero-shot generalization capabilities in authorship identification expanding their practical applicability.
  
  \item Empirical demonstration with ablations and challenges using the proposed architecture.
  
  \item Three case studies analyzing the behaviour of the model, supported by applications, including qualitative examination of the model.
  
\end{enumerate}
In the following sections we describe 
current approaches and state-of-the-art in language processing (Section~\ref{sec:related_work}), our implementation of the idea of authorship embeddings (Section~\ref{sec:ae}), the data used to train this model (Section~\ref{sec:data}) and results supporting the viability of this technique (Section~\ref{sec:exp}). A final discussion of these results is conducted in Section~\ref{sec:conclusions}, pointing to weaknesses and possible improvements.



\section{Related work} \label{sec:related_work}
    The online media content explosion has meant a paradigm shift in many areas of knowledge. For Natural Language Processing (NLP), billions of text pieces are available right now at an ever-increasing creation rate. The vast amount of authored texts allows us to apply modern state-of-the-art transformer models to the domain of authorship effectively, taking advantage of blogging pages or online libraries. In this section we determine authorship for the scope of our work and explore solutions that enable us to numerically represent this complex concept. 

    \subsection{Authorship}
        The act of writing is inherently influenced by the person performing it. Their beliefs, knowledge, mannerisms among others impact the contents of any given piece if the document is sufficiently extensive. Following this assumption, we determine authorship as a heterogeneous set of identifiable features of the text that uniquely belong to an author. Authorship attribution, for instance, tries to relate these stylistic features to the author~\cite{Stamatatos2009Mar}. Many areas of research have vested interest in determining authorship. For example linguistics, forensics or history focus on this task from different points of view~\cite{juola2008authorship}, and they could benefit from an accurate automated model of authorship. When attribution is considered, two features of text are typically used: content and the style~\cite{sundararajan2018represents}. Computers can easily identify manually extracted stylometric features such as word choice and frequency, punctuation or sentence length among others~\cite{sari2018topic}, building sets of representative characteristics that can be plugged into a machine learning approach such as an ensemble~\cite{custodio2018each}, aiming to discover differentiating features. 

        Applications of authorship attribution can proliferate in many domains, for example, in literature. In recent works, a LDA-Transformer neural network architecture was applied to Chinese poetry to identify authorship~\cite{ai2021lda}. Russian literacy is also treated~\cite{fedotova2022authorship}, with a comparison between several popular methods ranging from classical machine learning models fed with features to deep learning recurrent neural networks or transformers. More complex techniques can be applied to attribution, such is the case of Dynamic Authorship Attribution (DynAA)~\cite{custodio2021stacked}, where heterogeneous sources are merged with a stack of classifiers to improve the attribution of authorship.
        
        Generalization is a frequent concern in machine learning, and more so in the field of authorship attribution. Adaption to out-of-train domains or examples is usually hard and many methods fail to transfer to new datasets, as highlighted by Murauer et al.~\cite{murauer2021developing}. Including several datasets, with varying document size, other languages, genres and styles, is central to evaluate a model trained to solve authorship attribution task. There is also concern for domain bias, as shown by Hitschler et al.~\cite{hitschler2017authorship}. Together with these concerns, there are also challenges ahead~\cite{bevendorff2020overview}, such as the size of each author's document (i.e. a tweet) or the existing orthogonality between topics and authors. In case of large texts, other problems appear, such as co-authoring, which includes complex issues such as style change detection~\cite{deibel2021style}, or style changes~\cite{Nieuwazny2021Sep}.

        Embeddings provide a useful and rich representation of stylistic, semantic and syntactic properties that can reveal authorship. Recent approaches proposed in the literature include PolitiBETO~\cite{villa2022nlp}, a BETO~\cite{canete2020spanish} model trained on a corpus of political language and an ensemble approach profile authors in the same domain. Other authors have also used BETO, this time in combination with the model MarIA~\cite{gutierrez2022maria}. BERT~\cite{Devlin2018Oct} has focused a large number of works in the state-of-the-art literature. For instance, a solution based on the BERT model was proposed for a challenge called ``Profiling Irony and Stereotype Spreaders on Twitter''. Due to the limitation of this model to deal with large inputs, the authors propose a re-segmentation approach, concatenating all tweets belonging to the same author, then splitting the text to adapt to the expected entry of BERT~\cite{yu2022bert}. Other researchers propose the use of parallel stylometric document embeddings, using four different document embeddings models focused on classical information and an additional embedding generated with mBERT~\cite{math10050838}.
        
        Other authors also use BERT but with a siamese network~\cite{tyo2021siamese}, training the model aiming to generate close embeddings for text of the same author, while maximising the distance when authors are different. In this case, although the authors also use a triplet loss, they claim that there is no benefit. In another approach where BERT is used for authorship verification~\cite{futrzynski2021author}, the embeddings are extracted from short samples of texts, randomly sampled, and unified using the median. Other transformer-based model used for this task includes Sentence-BERT~\cite{reimers2019sentence}~\cite{schlicht2021unified}, employed with the goal of identifying hate speech spreaders. 

        Several researchers have also focused on assessing the quality of embeddings generated with different methods. For instance, Kumar et al.~\cite{kumar2022comparing} compare three classical embedding techniques, Doc2Vec, GloVE, and FastText. The results show that FastText CBOW reaches the best results, while GloVe requires more resources. However, the authors argument that there is no clear supremacy among the models tested. In another article, an evaluation framework for author embedding methods is presented~\cite{terreau2021writing}, focusing on those based on writing style, measuring the ability of a model to capture stylistic features, without considering the topic. It also presents an interesting conclusions, showing that recent models are highly influenced by inner semantic of authors' production. 

        Contributions of the state of the art have also studied different types of features to identify patterns of authorship. A research highlighted the importance of linguistic patterns to reveal authorship, proposing an approach focused on stylistic and semantic features and a T5 language model to provide embeddings, while a CNN and an attention mechanism provides local and contextual features~\cite{najafi2022text}. Focused on the identification of an author of an anonymized paper~\cite{10.1145/3018661.3018735} and to generate embeddings appropriate for author identification, the authors include a guidance process using a meta path selection, modelling anonymized papers as nodes in a bibliographic network and predicting neighbors of a node. DeepStyle also follows this research line~\cite{10.1007/978-3-030-60290-1_17}, an embedding framework focused on stylistic information, extracting salient features from the users' social media posts and a Triplet loss function. Other researchers highlight the importance of the syntactic structure of sentences to detect representative authorship patterns~\cite{10.1145/3491203}. The authors propose a semi-supervised framework for learning these characteristics, using a model that integrates a lexical and a syntactic sub-networks.
        
        After examining the literature we conclude that research focused on authorship is bounded by the data used to train the model. Models built with these techniques are successful at in-domain tasks, but fail at others. In our research we have found that using authored data, zero-shot classification of authors can be performed if, instead of classification, we adopt a contrastive approach. The next two sections explore our needs for this kind of training: \textit{pretrained Transformers}, which already produce very powerful representations (semantic and stylistic) and \textit{contrastive learning}, which allows a model to learn these representations.

    \subsection{Transformers and Representations}
        Transformers have become a staple of NLP in record time. Since the conception of Attention~\cite{Vaswani2017Jun} and the popularization of pretraining with BERT~\cite{Devlin2018Oct}, a plethora of out-of-the-box transformer models and architectures have been trained to understanding natural language. In turn, these models are able to interpret semantic features of speech, as well as some, albeit scarce, stylistic understanding.
        
        Works that focus on style are usually concerned with content generation rather than understanding, therefore they include a style vector into the system as in the Styleformer~\cite{Park2022}. Other modalities such as audio have seen certain attention to style~\cite{choi2020encoding} although without concern of authorship. Some interest has been put in an authorship transformer~\cite{Kalgutkar2019Feb} but for code writing.
        
        For our purposes, we require encoder transformers. Models such as RoBERTa~\cite{liu2019roberta} fit our purposes. Decoder-style models produce worse representations that their encoder counterparts, for example between BERT, ELMo~\cite{Peters2018Feb} and GPT-2~\cite{radford2019language}, the first usually produces more contextualized representations~\cite{ethayarajh2019contextual} leading scoreboards on similarity against their decoder counterparts. The closest transformer found to our task at hand is BertAA~\cite{fabien2020bertaa} where features and a transformer are combined to perform logistic regression to attribute authorship. The use of embeddings can help take authorship to a more complex level, enabling cross-domain scenarios with pre-trained models and a normalized corpus~\cite{barlas2020cross}. A particular scenario is the case for microtext from social media, which can be particularly challenging~\cite{suman2021authorship}. To generate embeddings we choose RoBERTa, as its representations have shown to be more robust than BERT-generated embeddings~\cite{liu2019roberta}. More sophisticated encoder models would be usable too at the cost of compute time and memory. 

    \subsection{Contrastive Learning}
        Recent advances show that, with enough amounts of data, self-supervised learning can achieve impressive generalization to zero-shot tasks. This is illustrated in works such as T5~\cite{Raffel2019Oct} or GPT-3~\cite{Brown2020May}, pre-trained with Causal Language Modeling (CLM) achieves state-of-the-art even in out-of-domain benchmarks. These models are extremely powerful, but semi-supervised training methods such as CLM or Masked Language Modeling (MLM) are semantically-oriented. A feasible solution is to learn a representation directly without masking or pseudo-labels via Contrastive Learning.
        
        Some works have started applying semi-supervised contrastive approaches to different domains to learn representations of text (and other modalities). This can be exemplified by CLIP~\cite{radford2021learning} where images and text are aligned for zero-shot classification, achieving outstanding performance in common benchmarks. Works like SimCSE~\cite{gao2021simcse} apply contrastive learning to sentence similarity tasks. Authorship has to be available for authorship attribution therefore labels can also be used to learn. Supervision can also be included to contrastive learning~\cite{khosla2020supervised} to improve the quality of representations. Such is the case for example for natural language inference~\cite{hu2022pair}, where results can be improved via Supervised Contrastive Learning (SCL) and then labelled. Relevant efforts have been recently made towards applying SCL to authorship representations \cite{rivera2021learning}, but transfer is still an open problem.
        
        To summarize our conclusions from this section, we will be using pre-trained transformers as they already have a semblance of stylometric capabilities, paired with a contrastive learning approach to build representations of an authors writing style.

\section{Authorship embeddings} \label{sec:ae}
We begin with an assumption: when an author writes a text, several features of speech are reflected throughout the work. Whether they are intentional (such as a literary author) or unintentional (from a social network user) they can be detected and measured~\cite{neal2017surveying}. If an expert were to hand-craft these features, he or she could identify abstract (and subjective) properties such as rhythm, punctuation, flow, registry, and so on. 
If our initial assumption is true and meaningful and representative features of speech can be extracted from diverse texts of an author, then it is expected the network to maximize the similarity of texts belonging to the same author. We name this representation of writing style authorship embedding. Whereas semantic embeddings generated by transformers detect contextual and semantic features focusing on the content of the text, authorship embeddings are meant to encode features (from context and semantics too) from the author of the text, shifting the focus of the transformer towards style.

\subsection{Problem definition}
We define authorship embeddings as a numerical representation of the writing style of an author. We want to approximate a function that takes a text and determines these numerical features. Let $D$ and $D'$ be document sets, where $D$ has a document per author and $D'$ has another different document per author in the exact same order as $D'$. From a set of documents $D = \{D_1, D_2, ..., D_N\}$ we find documents $D_i$ where $i$ represents the author identifier, we want to generate a corresponding embedding set $E = \{E_1, E_2, ..., E_N\}$ and $E'$ that represents each author writing style. Ideally, if the cosine similarity of $E$ and $E'$ was to be computed we would obtain an identity matrix as follows $S_c(E, E') = I_{NxN}$. We will follow this objective in the training procedure to achieve maximum similarity between embeddings from the same author.

Each embedding $E_i$ represent textual documents embedded in a numerical space, if the earlier objective is followed, will encode the features from the author by generating the same representation for different texts. This projection into a hyper-space allows to profile any author independently of the set of authors seen in the training set, as the space generated is a continuum of writing styles where each embedding is a point.

\subsection{Contrastive pretraining}
\begin{figure}[htbp]
\centerline{\includegraphics[width=0.35\textwidth]{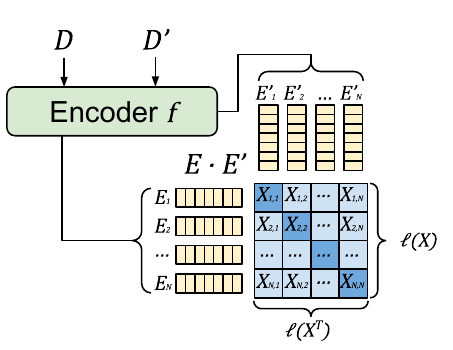}}
\caption{Visualization of InfoNCE loss computation for related document sets $D$ and $D'$.}
\label{fig:loss}
\end{figure}

To build a suitable function $f$ able to translate a document $D_i$ into its respective $E_i$ a proper loss function has to be found, in this case we opted to use InfoNCE~\cite{oord2018representation} loss modified to efficiently process batches of data in a way similar to CLIP. The loss process can be followed in Fig. \ref{fig:loss}, where the calculation is visualized.

To ensure learning of representations we have to build $D$ and $D'$ guaranteeing they follow some constraints. 
\begin{itemize}
    \item Every document in $D$ has to belong to a different author. Repeated authors in $D$ or $D'$ act as noise for the InfoNCE loss.
    \item Each author has a pool of documents to choose from, always larger than 2, although more documents are encouraged. When choosing documents $D_i$ and $D'_i$ for author $i$, both the anchor and positive examples are chosen at random. Other data augmentation methods aside from random sampling have been found to distort key textual features.
\end{itemize} 

After a suitable $D$ and $D'$ have been successfully built, the documents in both sets are transformed with an encoder architecture to build their respective embedding sets $E$ and $E'$ with such as $E_i = f(D_i)$. We want to compute the pairwise cosine similarity for the embedding sets, approximating $S_c(E, E') = I_{NxN}$. To compute $X$, the cosine similarity matrix, we perform Eq.~\ref{eq:logits}.

\begin{equation}
X = ||E||_2 \cdot ||E'^T||_2 \cdot \tau  \label{eq:logits}
\end{equation}

The term $\tau$ is a learnable temperature parameter tuned on-line during training, which is later used to compute the NT-Xent ~\cite{Sohn2016} loss over the similarity matrix as proposed in SimCLR~\cite{chen2020simple} and CLIP~\cite{radford2021learning}. Temperature regulates the L2 normalized product of embedding sets for the cross-entropy function as described in Eq.~\ref{eq:cce}.

\begin{equation}
\ell(X) = \frac{1}{N} \cdot \sum_{n=1}^N -\log \frac{\exp{X_{n,n}}}{\sum_{i=1}^N \exp{X_{n,i}}}  \label{eq:cce}
\end{equation}

The cross-entropy is applied to a square matrix, maximized when $X = I_{NxN}$ as our described objective. To ensure the efficient usage of data, the NT-Xent loss can also be computed on the transposed $X^T$ matrix, representing the similarity matrix $S_c(E', E)$, an efficient second pass as performed in Eq.~\ref{eq:lossfn}.

\begin{equation}
\mathcal{L}(X)= \frac{1}{2}(\ell(X) + \ell(X^T)) \label{eq:lossfn}
\end{equation}

\subsection{Network architecture}
An encoder architecture is needed to interpret the text and find the authorship embedding. First, we follow the RoBERTa-large architecture, tokenizing text into tokens of 512 chunks. The transformer has been frozen to preserve the transformer ability to interpret language as trained by the MLM loss. The semantic word embeddings are a matrix with dimension $(L, K)$ where $L$ is the sequence length and $K$ is the number of features.

Nonetheless, the semantic word embeddings have to be interpreted, and for this purpose we append a bidirectional LSTM to the architecture. The BiLSTM is more effective with lower amounts data points that a transformer layer, therefore the end representations are obtained with a recurrent network. We extract $K/2$ features for each LSTM pass, to form an embedding of size $K$. The authorship embedding is represented by this single vector of length $K$

\subsection{Hyper-parameters}
A summary of hyper-parameters used to pretrain our model are shared in Table \ref{tab:hyperparameters}. As the backbone frozen transformer we use RoBERTa-large~\cite{liu2019roberta}, as well as its tokenizer algorithm, using the public version currently available at the time of writing. We detail each hyper-parameter set for each dataset in Table~\ref{tab:hyperparameters}.

\begin{table}[]
\begin{center}
\caption{Table with hyperparameters used for training the network.}
\resizebox{0.4\textwidth}{!}{%
\begin{tabular}{@{}llll@{}}
\toprule
\textbf{Hyperparameter}    & \textbf{Books} & \textbf{Blogs} & \textbf{Mails} \\ \midrule
\textit{Batch size N}      & 2048           & 2048           & 64             \\
\textit{Optimizer}         & \multicolumn{3}{l}{Adam}                         \\
                           & \multicolumn{3}{l}{Training steps = 3000}        \\
                           & \multicolumn{3}{l}{Peak learning rate = 5e-3}    \\
                           & \multicolumn{3}{l}{Weight decay = 1e-2}          \\
                           & \multicolumn{3}{l}{Linear decay schedule}        \\
                           & \multicolumn{3}{l}{Momentum 1 = 0.9}             \\
                           & \multicolumn{3}{l}{Momentum 2 = 0.99}            \\
\textit{Sequence length L} & \multicolumn{3}{l}{512}                          \\
\textit{Embedding dim K}   & \multicolumn{3}{l}{1024}                         \\
\textit{Start temperature} & \multicolumn{3}{l}{exp(7e-2)}                    \\ \bottomrule
\end{tabular}%
}
\label{tab:hyperparameters}
\end{center}
\end{table}

Large batch sizes are usually required when computing InfoNCE loss, however the Mail dataset does not allow to use batch sizes larger than 64, which results in poor performance.

\section{Dataset} \label{sec:data}
High-quality datasets for authorship attribution are scarce, usually there is a need to adapt other datasets to this domain. Datasets that are specifically built for authorship are typically very small and end up making a very small amount of authors. Therefore, we consider and adapt three common benchmark datasets to build the model. Our pre-processing deviates from common approaches, therefore we contribute a general pre-processing pipeline and specific details for some datasets.

Our aim is to build authorship representations by comparing several pairs of authored texts simultaneously. Therefore, we require suitable pools of authored texts to pull from. Simple tuples of author and their text suffice, given that there are more than two texts belonging to the same author. Usually authored datasets contain a document and author, documents and authors are standardized as follows: 
\begin{enumerate}
    \item Each author-document set is merged with a separator </s> token.
    \item The resulting large text is pre-tokenized and split in chunks of 512 tokens.  
    \item Each chunk is then considered for training as long as the number of chunks for an individual author is 2 or more. 
    \item The dataset is split by authors. We train our model with 90\% of the authors and 10\% for training and testing respectively. We aim to evaluate the authors in the test set using the stylometric features learned in the training set without fine-tuning.
\end{enumerate}

We consider the following datasets for training and later validation:

\subsection{Standardized Gutenberg~\cite{gerlach2020standardized}}
This dataset contains authored books, as well as analyzable meta-data such as age or book type. Books with anonymous authors or authored by more than one person are discarded from our analysis. In the standardized corpus, there is some identifying information about the author at the beginning and end of the books. Therefore we opted to drop the first and last chunk of every book.

The Standardized Gutenberg corpus after clean-up and preprocessing has 635865 text chunks belonging to 1162 authors for training, with 53565 chunks from 129 authors in testing.
    
\subsection{Blog authorship~\cite{schler2006effects}}
This dataset contains blog posts from a diverse set of age groups. It is by far the largest dataset of the evaluation. Without further preprocessing, we extracted 343254 texts from 17287 authors for training and 38770 texts from additional 1920 authors for testing. 
    
\subsection{Enron mails~\cite{shetty2004enron}} 
This dataset contains emails from the Enron corporation, authored a hundred people. This is the most challenging dataset due to the small size and noisy nature of the emails. Email headers and footers are removed from the dataset as they contained identifying information about each author. There are 489843 texts by 135 accounts in training and 123792 texts from 15 accounts for testing.

\section{Experimentation} \label{sec:exp}
Validation of our model is difficult as there is limited literature focused on building representations for authorship. Therefore, we have designed an evaluation procedure for PART. To bridge this we split our experimentation in three sections, ablations for data usage and our authorship training, authorship challenges performed at zero-shot and with feature extraction, and three case studies on each of the testing datasets available.

\subsection{Baselines}
First we conduct ablations, we train several models removing the embeddings and test them on held-out parts of each dataset. Also we test the frozen RoBERTa encoder as out-of-the-box, to demonstrate the disentanglement from semantics and content. For challenges we experiment with RoBERTa to demonstrate the need for a pre-training, against DeBERTa-v3~\cite{He2021Nov}, to show that a more powerful model does not equal better authorship attribution. Finally we apply the Style Embeddings~\cite{wegmann-etal-2022-author}, which uses a similar technique with contrastive learning.

\subsection{Ablations on attribution}
We evaluate the accuracy of our model at zero-shot attribution of 512 tokens from a reference author from a pool of $N-1$ other authors. Accuracy is shown on Table \ref{tab:global-zero}
. We refer to PART as the model trained with the full combination of datasets. Accuracy is measured as the average of 100 rounds of trials, with the standard deviation contiguous to the accuracy value. A trial is defined as follows: 1) Pick an author at random 2) Pick two chunks from the same author at random 3) repeat 1 and 2 $N$ times without repetition 4) compute similarity and performance metrics. We list models trained on different datasets, including a combination of all authors from all datasets.

\begin{table*}[]
\resizebox{\textwidth}{!}{%
\begin{tabular}{@{}lll|llll@{}}
\toprule
Model                         & N=                    & Metric         & Blogs+Books+Mails & Blogs           & Books           & Mails           \\ \midrule
\multirow{10}{*}{\begin{tabular}[c]{@{}l@{}}PART model\end{tabular}} &
  \multirow{2}{*}{=10} &
  Accuracy &
  91.25\% ± 7.69 &
  90.60\% ± 8.90 &
  87.75\% ± 10.08 &
  28.65\% ± 9.79 \\
                              &                       & Top-5 Accuracy & 98.35\% ± 3.47    & 98.25\% ± 4.02  & 98.85\% ± 3.31  & 68.85\% ± 10.02 \\
                              & \multirow{2}{*}{=20}  & Accuracy       & 86.60\% ± 7.16    & 88.08\% ± 6.68  & 81.45\% ± 7.21  & -               \\
                              &                       & Top-5 Accuracy & 95.40\% ± 4.28    & 96.70\% ± 3.59  & 96.28\% ± 3.86  & -               \\
                              & \multirow{2}{*}{=50}  & Accuracy       & 83.72\% ± 4.60    & 82.51\% ± 4.78  & 72.14\% ± 5.34  & -               \\
                              &                       & Top-5 Accuracy & 93.73\% ± 3.54    & 93.13\% ± 3.58  & 91.94\% ± 3.26  & -               \\
                              & \multirow{2}{*}{=100} & Accuracy       & 78.39\% ± 3.99    & 77.95\% ± 3.95  & 65.07\% ± 3.89  & -               \\
                              &                       & Top-5 Accuracy & 90.84\% ± 2.55    & 90.14\% ± 2.90  & 86.49\% ± 2.81  & -               \\
                              & \multirow{2}{*}{=250} & Accuracy       & 72.39\% ± 2.39    & 71.56\% ± 2.45  & -               & -               \\
                              &                       & Top-5 Accuracy & 86.73\% ± 2.07    & 85.77\% ± 1.92  & -               & -               \\ \midrule
\multirow{10}{*}{Blogs model} & \multirow{2}{*}{=10}  & Accuracy       & 90.35\% ± 8.67    & 91.55\% ± 8.21  & 64.15\% ± 12.71 & 26.25\% ± 11.76 \\
                              &                       & Top-5 Accuracy & 97.55\% ± 4.50    & 97.75\% ± 4.49  & 92.75\% ± 6.72  & 67.65\% ± 13.52 \\
                              & \multirow{2}{*}{=20}  & Accuracy       & 86.67\% ± 7.81    & 87.68\% ± 7.95  & 56.17\% ± 9.49  & -               \\
                              &                       & Top-5 Accuracy & 96.10\% ± 3.99    & 96.05\% ± 3.97  & 85.32\% ± 6.56  & -               \\
                              & \multirow{2}{*}{=50}  & Accuracy       & 82.22\% ± 5.16    & 82.07\% ± 4.16  & 43.74\% ± 5.90  & -               \\
                              &                       & Top-5 Accuracy & 93.16\% ± 3.60    & 92.99\% ± 3.19  & 70.78\% ± 5.69  & -               \\
                              & \multirow{2}{*}{=100} & Accuracy       & 77.37\% ± 3.94    & 77.73\% ± 3.86  & 35.94\% ± 3.69  & -               \\
                              &                       & Top-5 Accuracy & 90.39\% ± 2.99    & 90.00\% ± 2.89  & 60.44\% ± 4.05  & -               \\
                              & \multirow{2}{*}{=250} & Accuracy       & 70.51\% ± 2.74    & 71.11\% ± 2.78  & -               & -               \\
                              &                       & Top-5 Accuracy & 85.65\% ± 2.19    & 85.50\% ± 2.20  & -               & -               \\ \midrule
\multirow{10}{*}{Books model} & \multirow{2}{*}{=10}  & Accuracy       & 58.25\% ± 13.16   & 52.60\% ± 13.59 & 83.50\% ± 11.15 & 21.80\% ± 10.62 \\
                              &                       & Top-5 Accuracy & 88.85\% ± 9.38    & 87.35\% ± 8.84  & 96.75\% ± 4.82  & 63.40\% ± 11.51 \\
                              & \multirow{2}{*}{=20}  & Accuracy       & 45.98\% ± 11.06   & 41.62\% ± 9.77  & 77.92\% ± 8.70  & -               \\
                              &                       & Top-5 Accuracy & 76.40\% ± 8.26    & 73.10\% ± 9.04  & 93.40\% ± 4.70  & -               \\
                              & \multirow{2}{*}{=50}  & Accuracy       & 34.93\% ± 6.33    & 30.62\% ± 5.84  & 70.10\% ± 6.62  & -               \\
                              &                       & Top-5 Accuracy & 62.19\% ± 5.96    & 57.20\% ± 6.55  & 89.28\% ± 3.78  & -               \\
                              & \multirow{2}{*}{=100} & Accuracy       & 28.06\% ± 4.26    & 24.88\% ± 3.35  & 61.38\% ± 4.34  & -               \\
                              &                       & Top-5 Accuracy & 50.18\% ± 4.94    & 47.44\% ± 4.13  & 83.51\% ± 2.92  & -               \\
                              & \multirow{2}{*}{=250} & Accuracy       & 20.90\% ± 2.09    & 17.05\% ± 1.96  & -               & -               \\
                              &                       & Top-5 Accuracy & 38.08\% ± 2.73    & 33.81\% ± 2.44  & -               & -               \\ \midrule
\multirow{10}{*}{Mails model} & \multirow{2}{*}{=10}  & Accuracy       & 25.40\% ± 11.48   & 21.30\% ± 10.38 & 26.90\% ± 12.22 & 24.90\% ± 12.43 \\
                              &                       & Top-5 Accuracy & 70.40\% ± 14.10   & 68.10\% ± 14.40 & 70.55\% ± 10.84 & 68.80\% ± 12.65 \\
                              & \multirow{2}{*}{=20}  & Accuracy       & 18.40\% ± 8.76    & 13.77\% ± 6.26  & 16.93\% ± 7.38  & -               \\
                              &                       & Top-5 Accuracy & 47.82\% ± 11.84   & 42.43\% ± 9.20  & 48.10\% ± 9.37  & -               \\
                              & \multirow{2}{*}{=50}  & Accuracy       & 10.47\% ± 4.12    & 8.56\% ± 3.42   & 8.94\% ± 3.49   & -               \\
                              &                       & Top-5 Accuracy & 28.72\% ± 6.13    & 24.13\% ± 5.72  & 27.94\% ± 4.98  & -               \\
                              & \multirow{2}{*}{=100} & Accuracy       & 6.84\% ± 1.87     & 5.47\% ± 2.13   & 6.29\% ± 2.05   & -               \\
                              &                       & Top-5 Accuracy & 19.39\% ± 3.83    & 15.86\% ± 3.45  & 18.55\% ± 3.64  & -               \\
                              & \multirow{2}{*}{=250} & Accuracy       & 4.27\% ± 1.06     & 3.05\% ± 0.89   & -               & -               \\
                              &                       & Top-5 Accuracy & 11.87\% ± 1.69    & 8.95\% ± 1.53   & -               & -               \\ \midrule
\multirow{10}{*}{RoBERTa Baseline} &
  \multirow{2}{*}{=10} &
  Accuracy &
  44.65\% ± 15.02 &
  39.20\% ± 13.61 &
  44.05\% ± 13.98 &
  23.85\% ± 10.46 \\
                              &                       & Top-5 Accuracy & 76.55\% ± 11.24   & 72.50\% ± 9.39  & 77.90\% ± 10.82 & 63.00\% ± 11.45 \\
                              & \multirow{2}{*}{=20}  & Accuracy       & 36.95\% ± 10.37   & 34.42\% ± 8.93  & 37.15\% ± 9.89  & -               \\
                              &                       & Top-5 Accuracy & 61.33\% ± 9.41    & 58.30\% ± 8.56  & 63.62\% ± 8.03  & -               \\
                              & \multirow{2}{*}{=50}  & Accuracy       & 27.65\% ± 5.61    & 24.53\% ± 4.80  & 28.91\% ± 6.14  & -               \\
                              &                       & Top-5 Accuracy & 45.89\% ± 6.49    & 41.34\% ± 6.19  & 50.39\% ± 6.47  & -               \\
                              & \multirow{2}{*}{=100} & Accuracy       & 23.30\% ± 4.01    & 21.30\% ± 3.82  & 24.23\% ± 3.31  & -               \\
                              &                       & Top-5 Accuracy & 38.09\% ± 4.46    & 34.96\% ± 4.17  & 42.11\% ± 3.73  & -               \\
                              & \multirow{2}{*}{=250} & Accuracy       & 18.77\% ± 1.99    & 16.76\% ± 1.86  & -             & -               \\
                              &                       & Top-5 Accuracy & 30.03\% ± 2.29    & 26.94\% ± 2.49  & -               & -               \\ \bottomrule
\end{tabular}%
}
\caption{Zero shot accuracy and top-5 accuracy for our method. N represents the number of authors picked at random to create a document and its reference, the higher the N the difficulty of finding the referenced author increases. If there are less authors in a testing dataset than N, the space is left blank.}
\label{tab:global-zero}
\end{table*}

To begin with, we observe that the RoBERTa baseline performs below most models, except the Mails model. The baseline, despite being untrained, holds fair scores across tasks, performing much better than random chance. The Mails model has been trained with a very low batch size, which explains its under-par performance. However, every other model outperforms the RoBERTa baseline on most cases. For instance, the Books model has better out-of-domain performance on the blogs dataset than the baseline, while the same is true for the Blogs model on the books dataset. All models perform best when trained with in-domain data. 

On every metric, the model trained on the combination of all datasets outperforms every other trained model. It achieves $72.39\%$ accuracy and $86.63\%$ top-5 accuracy on a set of 250 documents. Observed accuracy values are consistently high on the combined domain, with averages ranging from $91\%$ to $72\%$, and top-5 scores within $98\%$ to $86\%$. Observation makes clear that increasing the document set $N$ makes the task significantly harder, but with a more consistent accuracy. Interestingly, the combined model, has top performance on all datasets, including the book dataset performance which is boosted by the inclusion of additional mail and blog data. Gains on the blogs dataset are marginal and non-significant, varying between document sets with high standard deviation.
\begin{table*}[htpb!]
\caption{PAN challenges from 2012 to 2019 on Authorship attribution.}

\resizebox{\textwidth}{!}{%

\begin{tabular}{@{}l|llll|llll@{}}
\toprule
                      & \multicolumn{4}{c|}{\textbf{Accuracy}}                                                                                         & \multicolumn{4}{c}{\textbf{F1-Score}}                                                                                         \\
Challenge      & PART              & DeBERTa-v3 & RoBERTa  & Style-Embedding & PART              & DeBERTa-v3 & RoBERTa  & Style-Embedding \\ \midrule
PAN 11 - Large        & \textbf{37,54 \%}        & 13,00 \%                       & 28,69 \%                    & 25,08 \%                             & \textbf{26,18 \%}        & 8,15 \%                        & 18,82 \%                    & 16,47 \%                            \\
PAN 12 - A            & \textbf{83,33 \%}        & 66,67 \%                       & 50,00 \%                    & \textbf{100,00 \%}                   & \textbf{82,22 \%}        & 65,56 \%                       & 41,27 \%                    & 100,00 \%                           \\
PAN 12 - B            & \textbf{100,00 \%}       & 66,67 \%                       & 66,67 \%                    & 66,67 \%                             & \textbf{100,00 \%}       & 65,56 \%                       & 65,56 \%                    & 55,56 \%                            \\
PAN 12 - C            & \textbf{100,00 \%}       & 62,50 \%                       & 100,00 \%                   & 62,50 \%                             & \textbf{100,00 \%}       & 58,33 \%                       & 100,00 \%                   & 58,33 \%                            \\
PAN 12 - D            & \textbf{100,00 \%}       & 25,00 \%                       & 62,50 \%                    & 75,00 \%                             & \textbf{100,00 \%}       & 16,67 \%                       & 54,17 \%                    & 68,75 \%                            \\
PAN 12 - I            & 57,14 \%                 & 50,00 \%                       & \textbf{64,29 \%}           & 50,00 \%                             & \textbf{53,57 \%}        & 45,24 \%                       & \textbf{53,57 \%}           & 41,67 \%                            \\
PAN 12 - J            & 64,29 \%                 & 64,29 \%                       & \textbf{71,43 \%}           & 42,86 \%                             & 57,14 \%                 & 57,14 \%                       & \textbf{66,67 \%}           & 36,90 \%                            \\
PAN 18 - 20 authors    & \textbf{62,03 \%}        & 27,85 \%                       & 46,84 \%                    & 49,37 \%                             & \textbf{55,49 \%}        & 23,08 \%                       & 33,59 \%                    & 41,31 \%                            \\
PAN 18 - 15 authors    & \textbf{63,51 \%}        & 28,38 \%                       & 47,30 \%                    & 52,70 \%                             & \textbf{52,20 \%}        & 23,23 \%                       & 31,86 \%                    & 42,88 \%                            \\
PAN 18 - 10 authors    & \textbf{82,50 \%}        & 40,00 \%                       & 62,50 \%                    & 62,50 \%                             & \textbf{75,44 \%}        & 34,82 \%                       & 56,17 \%                    & 50,29 \%                            \\
PAN 18 - 5 authors     & \textbf{68,75 \%}        & 43,75 \%                       & \textbf{68,75 \%}           & 62,50 \%                             & 54,55 \%                 & 47,27 \%                       & 50,73 \%                    & \textbf{54,85 \%}                   \\
PAN 19 - r = 100\%     & \textbf{85,04 \%}        & 59,62 \%                       & 73,50 \%                    & 82,48 \%                             & \textbf{71,66 \%}        & 41,46 \%                       & 57,30 \%                    & 68,43 \%                            \\
PAN 19 - r = 80\%      & \textbf{48,98 \%}        & 25,51 \%                       & 46,94 \%                    & 51,02 \%                             & 39,86 \%                 & 24,93 \%                       & 40,36 \%                    & \textbf{47,02 \%}                   \\
PAN 19 - r = 60\%      & \textbf{53,79 \%}        & 28,03 \%                       & 50,00 \%                    & 49,24 \%                             & 47,14 \%                 & 27,01 \%                       & 46,65 \%                    & \textbf{47,29 \%}                   \\
PAN 19 - r = 40\%      & \textbf{48,68 \%}        & 48,68 \%                       & 16,45 \%                    & 85,53 \%                             & \textbf{60,35 \%}        & 33,59 \%                       & 27,53 \%                    & 43,27 \%                            \\
PAN 19 - r = 20\%      & \textbf{71,97 \%}        & 41,67 \%                       & 71,21 \%                    & 42,42 \%                             & \textbf{54,20 \%}        & 27,08 \%                       & 50,67 \%                    & 43,27 \%                            \\ \bottomrule
\end{tabular}
}
\label{tab:pan-attr-over}

\end{table*}

The mails dataset presents very low accuracies overall not higher than 28\%, being unable to be profiled by any model. We further explore this dataset in Section \ref{sub:enron} to analyse this behaviour.

\subsection{Authorship in challenges}
The earlier section was close to the domain of application, we perform experiments on common shared task from the PAN@CLEF. We experiment on two sets of tasks, a first one where the task is identical to the training with a new unseen domain, performing authorship attribution at zero-shot. The second set is style-change detection, which is a different task from the original training, if our claims are sound, PART should behave well as a frozen feature extractor which will lead to better results in the given competition.
\begin{itemize}
\item PAN@CLEF 11 authorship attribution~\cite{argamon_shlomo_2011_3713246}: A dataset of authorship attribution using short-texts.
\item PAN@CLEF 12 authorship attribution~\cite{juola_patrick_2012_3713273}: Diverse set of problems of traditional authorship attribution.
\item PAN@CLEF 18 cross-domain authorship attribution~\cite{kestemont_mike_2018_3737849}: Multilingual dataset of texts for attribution.
\item PAN@CLEF 19 cross-domain authorship attribution~\cite{kestemont_mike_2019_3530313}: For the purpose of this paper, similar to the 18' dataset.
\item PAN@CLEF 20-22 style change detection~\cite{zangerle2020overview, bevendorff2021overview, bevendorff2022overview} are datasets used for style change detection, detecting documents single or multi-authored, the number of authors or if there has been a style change between paragraphs. The task we approach is to detect where a style switch has happened in a text, at a paragraph level.
\item PAN@CLEF 23 multi-author writing style analysis~\cite{bevendorff2023overview} as with other editions, the goal is to find changes. However, at the time of writing, the challenge is ongoing and there is no leaderboard to compare to.

\end{itemize}

These datasets share a common relevant feature for our model, and is that the target text is dissimilar from the original training domain. With this evaluation we want to inform of the model capability to adapt at zero-shot to new styles from twitter messages to amateur fanfiction; the contribution of the contrastive pre-training to the transformer. To this end we evaluate on two tasks, authorship attribution and style change detection.

\begin{table*}[htpb!]
\caption{PAN challenges from 2011 to 2019 on Authorship attribution, compared against the public leaderboard.}

\resizebox{\textwidth}{!}{%

\begin{tabular}{@{}l|llll|llll@{}}
\toprule
               & \multicolumn{4}{c|}{\textbf{Accuracy}}                      & \multicolumn{4}{c}{\textbf{F1-Score}}                       \\ 
Challenge      & PART              & DeBERTa-v3 & RoBERTa  & Style-Embedding & PART              & DeBERTa-v3 & RoBERTa  & Style-Embedding \\ \midrule
PAN 11 - Large & \textbf{37,54 \%} & 13,00 \%   & 28,69 \% & 25,08 \%        & \textbf{26,18 \%} & 8,15 \%    & 18,82 \% & 16,47 \%        \\
PAN 12         & \textbf{84,13 \%} & 55,85 \%   & 69,15 \% & 66,17 \%        & \textbf{82,16 \%} & 51,42 \%   & 63,54 \% & 60,20 \%        \\
PAN 18         & \textbf{69,20 \%} & 34,99 \%   & 56,35 \% & 56,77 \%        & \textbf{59,42 \%} & 32,10 \%   & 43,09 \% & 47,33 \%        \\
PAN 19         & \textbf{61,69 \%} & 40,70 \%   & 51,62 \% & 62,14 \%        & \textbf{54,64 \%} & 30,81 \%   & 44,50 \% & 49,86 \%        \\ \bottomrule
\end{tabular}
}
\label{tab:pan-attr-comp}

\end{table*}

Attribution is performed without further tuning of the model, transforming texts into embeddings. Each 512 chunk of tokens produces a single embedding, every chunk belonging to an author is considered a document. If an unknown document matches with any of the candidate documents, it is considered from the same author. As our intent is to employ the model as close to zero-shot as possible, we approach style-detection challenges with minimal adaption, training a classification feedforward network from the embeddings. Thus the model is used exclusively for feature extraction, using the authorship embedding together with a classifier that is trained after the embeddings are obtained. 

First we show the contribution of the pre-training in Table~\ref{tab:pan-attr-over}. In short, we find that, for most problems and tasks, PART has been able to correctly attribute more documents to their owners than the baselines. Style-Embeddings do not achieve significant improvements over part except on the last challenge. DeBERTa-v3 is vastly inferior to both PART and all other baselines, showing that better semantic embeddings from pre-trained encoders do not translate well to our task, also showing that upscaling to better encoders does not guarantee performance improvements. RoBERTa has fair performance at zero-shot but is still inferior to itself and the pretraining as it would be expected.

On average we succeed on improving upon every task against the considered techniques as shown in Table~\ref{tab:pan-attr-comp}. PART is, thus, strictly superior at out-of-domain authorship attribution than other tested methods. If our results are compared to those in the competition, PART still lags behind n-gram models and other manual extraction methods, but remains competitive keeping upin the top-5 for most challenges. Interestingly, as late as 2019, no models in the competitions employ Transformers or neural networks, all rely on ngrams (and or features) and a classifier; also in 2019, the baseline IMPOSTERS is the worst performing in the ranking which uses a distance-based approach similar to ours.

We also perform experiments on style change detection to determine the effectiveness on our method on a similar task, which are reported in Table~\ref{tab:pan-style-change}. In short, all results from PART outperform the baselines with an exception on the PAN 2023 dataset 3 against style embeddings. We point out the achieved stability by PART, for example content-based methods fail at PAN 2023, 2022 (datasets 2 and 3) and 2021 while style embeddings fail in the 2020 dataset. To PART this is indifferent, being able to achieve high performance on either dataset.

\begin{table}[htpb!]
\caption{PAN challenges from 2020 to 2023 on Style change detection F1-scores. Best scores in bold. }

\resizebox{\linewidth}{!}{%
\begin{tabularx}{1.3\linewidth}{@{}lXXXX@{}} 
\toprule
                          & PART             & DeBERTa-v3       & RoBERTa & Style-Embedding  \\ \midrule
PAN 2023 Dataset 1        & \textbf{86,69\%} & 80,76\%          & 80,47\% & 70,60\%          \\
PAN 2023 Dataset 2        & \textbf{74,39\%} & 60,36\%          & 41,33\% & 73,80\%          \\
PAN 2023 Dataset 3        & 62,21\%          & 52,54\%          & 36,15\% & \textbf{64,62\%} \\
PAN 2022 Task 1 Dataset 1 & \textbf{79,51\%} & \textbf{79,51\%} & \textbf{79,51\%} & 59,06\%          \\
PAN 2022 Task 1 Dataset 2 & \textbf{69,32\%} & 55,47\%          & 48,02\% & 68,02\%          \\
PAN 2022 Task 1 Dataset 3 & \textbf{62,27\%} & 53,30\%          & 40,13\% & 61,40\%          \\
PAN 2021 Task 2         & \textbf{69,42\%} & 48,57\%          & 40,13\% & 69,33\%          \\
PAN 2020 Task 2 (narrow)    & \textbf{84,12\%} & 83,51\%          & 82,71\% & 59,82\%          \\
PAN 2020 Task 2 (wide)      & \textbf{82,38\%} & 77,35\%          & 75,41\% & 61,22\%          \\ \bottomrule
\end{tabularx}
}
\label{tab:pan-style-change}

\end{table}

When compared agains another feature extractors, either for content or style, PART dominates both in zero-shot and as a feature extractor. When compared with results in the competition, PART remains competitive against state of the art methods, including ensembles and other dominant approaches.






\begin{figure}[htbp!]
\centerline{\includegraphics[width=1\linewidth]{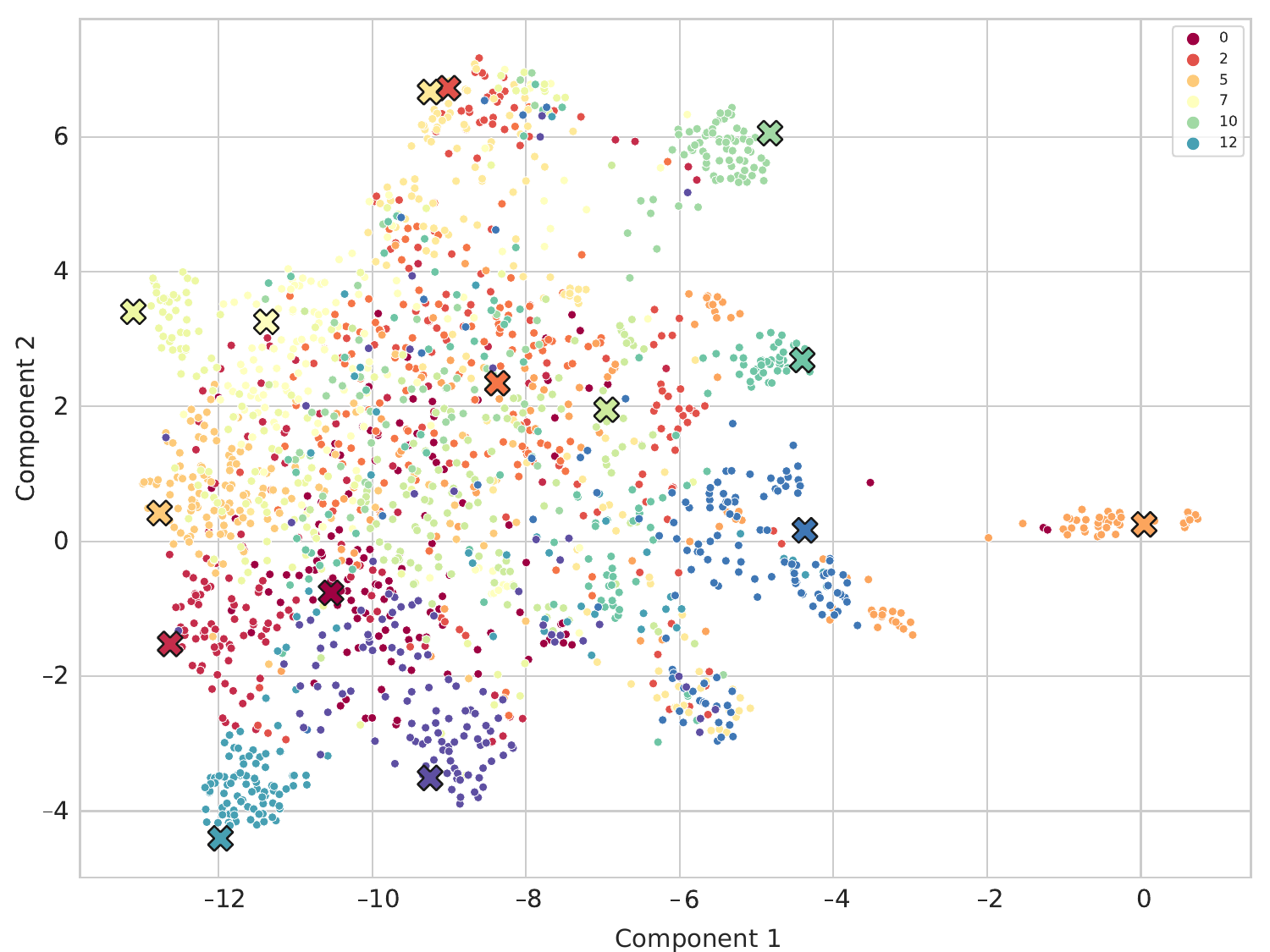}}
\caption{U-map representation of mails sent in the Enron dataset. Average representation marked with a cross, acting as a centroid. Parameters were 25 neighbors and 0.5 minimum cosine distance.}
\label{fig:mails}
\end{figure}

\subsection{Case study 1: Enron mail dataset} \label{sub:enron}
Two out of three datasets are easier to analyse with our method. We delve in the Enron email dataset to find justifications for the low performance of the representation model. To examine the quality of the representations we apply u-map, adjusting the hyperparameters for generating a 2d projection of features. We represent each mail as points in a plane, where the average of all representations of an author acts as a centroid.

\subsubsection{Authors and centroids}

In figure \ref{fig:mails} we projected the test split of the mailing dataset embeddings for visualization. First we point out that embeddings are grouped around the author centroid with some separation. Some groups form differentiable clusters, for example the rightmost group or the top-right (10th author) group. There are examples of other well represented authors with some overlap, such as the bottom-left groups. Here we observe groups of messages with moderate overlap, where differences are harder to determine but still end up closer to their centroid. Finally, the center of the representation heavily overlaps text points and, where our representations mostly fail to group embeddings around a centroid.

Sections in the hyperspace may overlap for separable authors, while still correctly grouping each text chunk around a cluster. The method is optimizing the target space to maximize authorship similarity. On the other hand, this method does not distribute the space evenly, as presented by the leftmost cluster. Some authors may be so distinct they occupy an entire region or just a separate area from the whole dataset. The projection of Fig.~\ref{fig:mails} has allowed frame the failings of the zero-shot attribution. Using previous observations we conclude that the representations need to be very powerful for zero-shot classification, but that is not as required for characterization and analysis of these embeddings.

\subsection{Case study 2: The Gutenberg corpus}
We evaluate the ability of our model to represent entire documents, using literature and books as a pretext. Each chunk produces an authorship embedding and averaging all chunks gives us the global authorship embedding for an author or book. This representation can be related to other books and to the book type, we achieve the first with a graph representation and the second with another u-map projection. Also we are interested in finding what each feature represents and what information does the embedding contain, as such we find the correlation between genre and individual feature.

\subsubsection{Book representations and authors}
\begin{figure*}[htbp!]
\centerline{\includegraphics[width=0.7\textwidth]{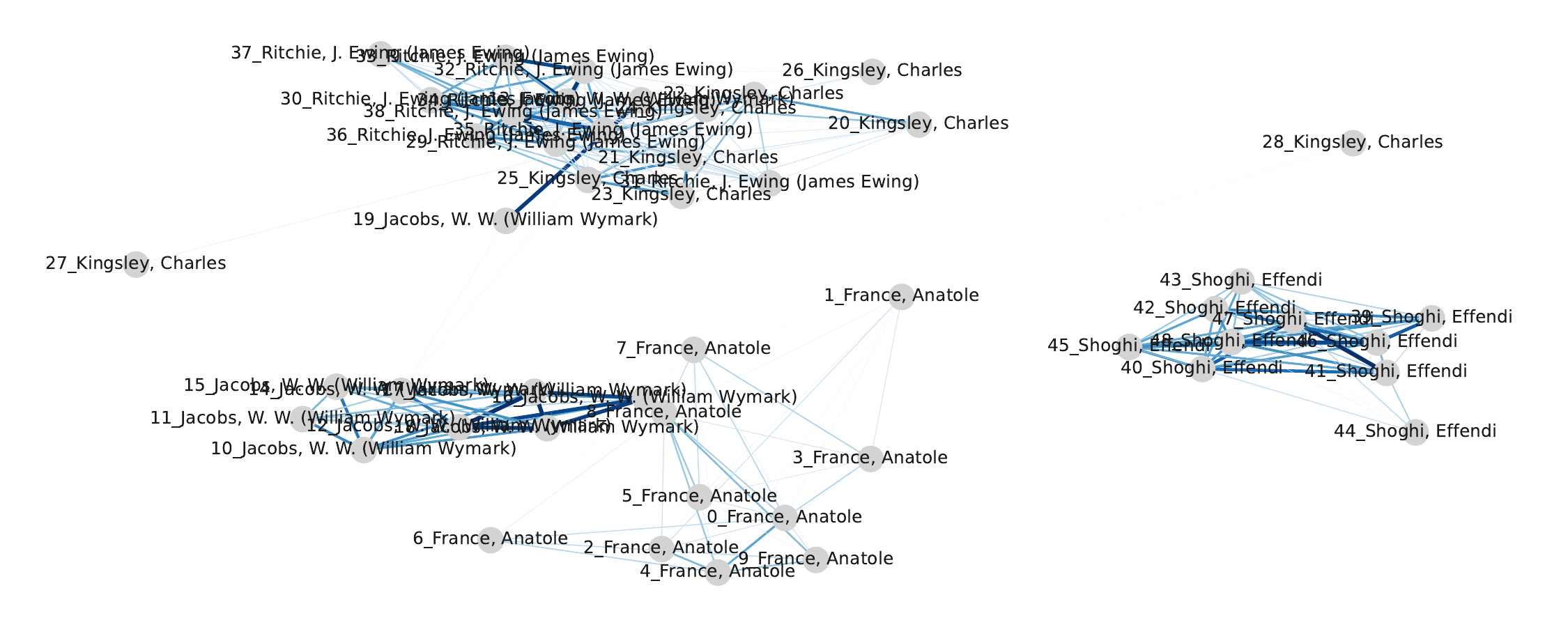}}
\caption{Graph of books, where nodes are books (with an author annotation) and edges represent presence of similarity above a threshold ($0.6$). Wider edges represent stronger similarity.}
\label{fig:graphbooks}
\end{figure*}

Fig. \ref{fig:graphbooks} shows the graph representation. It is constructed with books as nodes and edges as similarity above a threshold ($0.6$) of cosine similarity, where wider edges have stronger similarity. We have selected the 5 more represented authors with 10 of their works. A quick glance indicates that authors form clusters in this graph, with Shoghi and Jacobs tightly knit, France connections are weaker but also form a cluster and finally there is a mixed cluster of Ritchie, Kingsley and a work from Jacobs. The presence of this mix points to similar writing styles. Both began writing at the same period and shared interests on historical and political works, so a degree of similarity could be expected.

\subsubsection{Writing style and genre}
\begin{figure}[htbp]
\centerline{\includegraphics[width=0.8\linewidth]{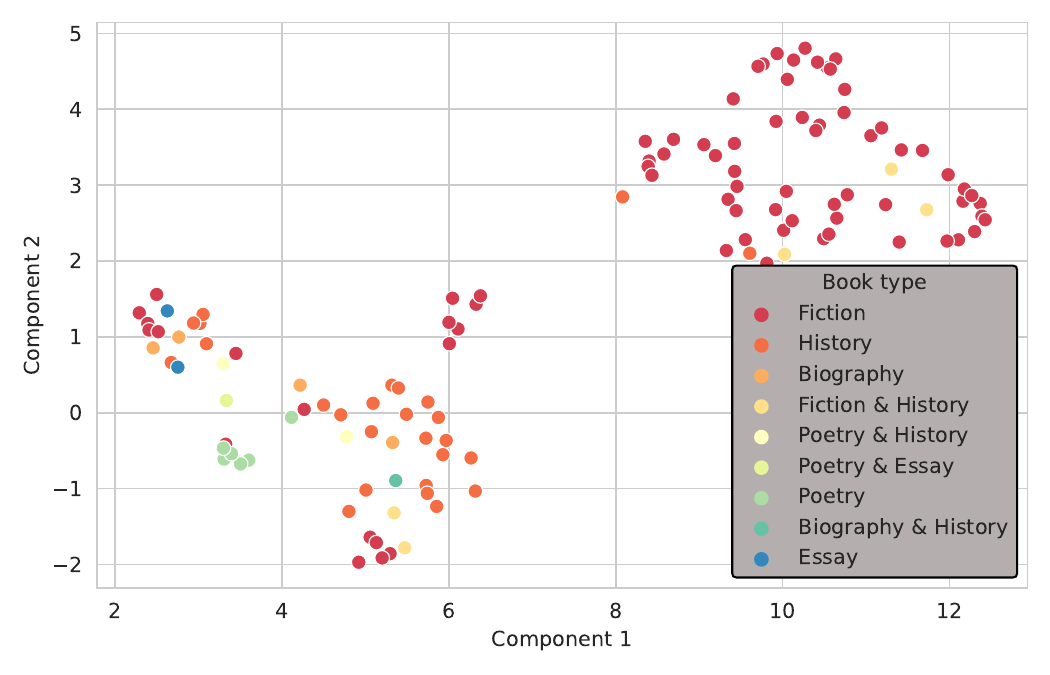}}
\caption{U-map representation of books. Legend represents the type of book, as determined by the Gutenberg dataset. Parameters were 10 neighbors and 0.1 minimum cosine distance.}
\label{fig:graphumap}
\end{figure}

The similarity between books with styles is further explored in figure \ref{fig:graphumap}. Here the book embeddings are projected with u-map. First we point out the top-right group of books, which are fiction and historical fiction. Similarity is closer for books of this style, as shown by the representation, historical fiction is more similar to most fiction than it is to history books. On the other hand there are some poetry books in the dataset with very distinct style, very similar to each other in the bottom-down region of the space. Also history books all share the same region of the figure. Some fiction is mixed with history books, and essays are close to historical books and fiction too. This overlap may be due to very similar styles or underrepresented essays in training. The model also finds similarity between Biographies and History books, which is to be expected as they share similar writing patterns. The projection shows most works are correctly grouped in a clearly-bounded region when the type of book is evaluated.

This is a side-effect to determining the writing style of an author, as most careers usually focus on a single type of writing. We aim to find correlation between the generated features and the type of book analysed. 

In contrast to the results obtained in the Mailing dataset, we find the results for the Gutenberg corpus much more precise. This accuracy impacts the quality of the embeddings and thus we can easily build full-document embeddings. This approach is usual in the NLP domain and, as the graph points out, associations are frequently correct when determining clusters and similar authors. The model has also shown that this training has the side effect of book type recognition, which is outside our initial scope. Despite the usefulness of book type recognition, it may pose undesired consequences; for example assume an author writes both poetry and prose, their books could be misrepresented due to the differences between both styles.

\subsection{Case study 3: Blog authorship dataset} 
Finally, we evaluate the ability of the joined model on the Blog authorship corpus. This is composed of blog posts and conveniently annotated with age, gender and occupation. This enables our features to relate to these annotations, allowing for close examination of relationships between features and authors. We study demographic information (age, gender and occupation) as annotated by the dataset. As before, we find correlation and make projections. Instead of analysing by chunk of text or blog post, we also find centroids for some of these annotations.   


\subsubsection{Demographic representation}

\begin{figure}[htbp]
\centerline{\includegraphics[width=0.95\linewidth]{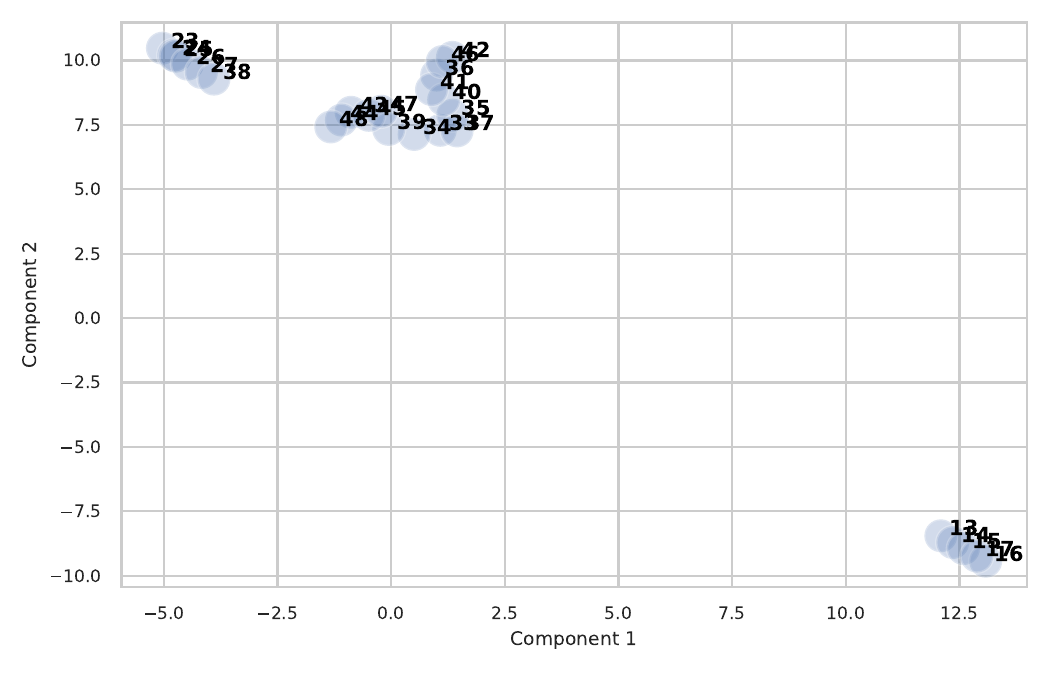}}
\caption{U-map projection of the combined embeddings by age group. Parameters were 3 neighbors and 0.5 minimum cosine distance.}
\label{fig:umapage}
\end{figure}

We examine the age of writers. There are a lot of ages in the dataset, distributed in groups (13-17; 22-27; over 34). We aggregate chunks of texts from all authors sharing the same age. These aggregates are projected with u-map in Fig.~\ref{fig:umapage}. The results obtained relate exactly to the aforementioned age groups except for the 38th point which is an outlier. There are three distinct groups, one for teenagers in the bottom right, other for young people in the top left corner and a cluster of middle-age people at the side of the young people group. Surprisingly, teenagers are sequentially ordered except for 16 and 17 years which are swapped. Young people have sequential order too, excluding the 38 years old representation.

\subsubsection{Occupation}
\begin{figure}[htbp]
\centerline{\includegraphics[width=1\linewidth]{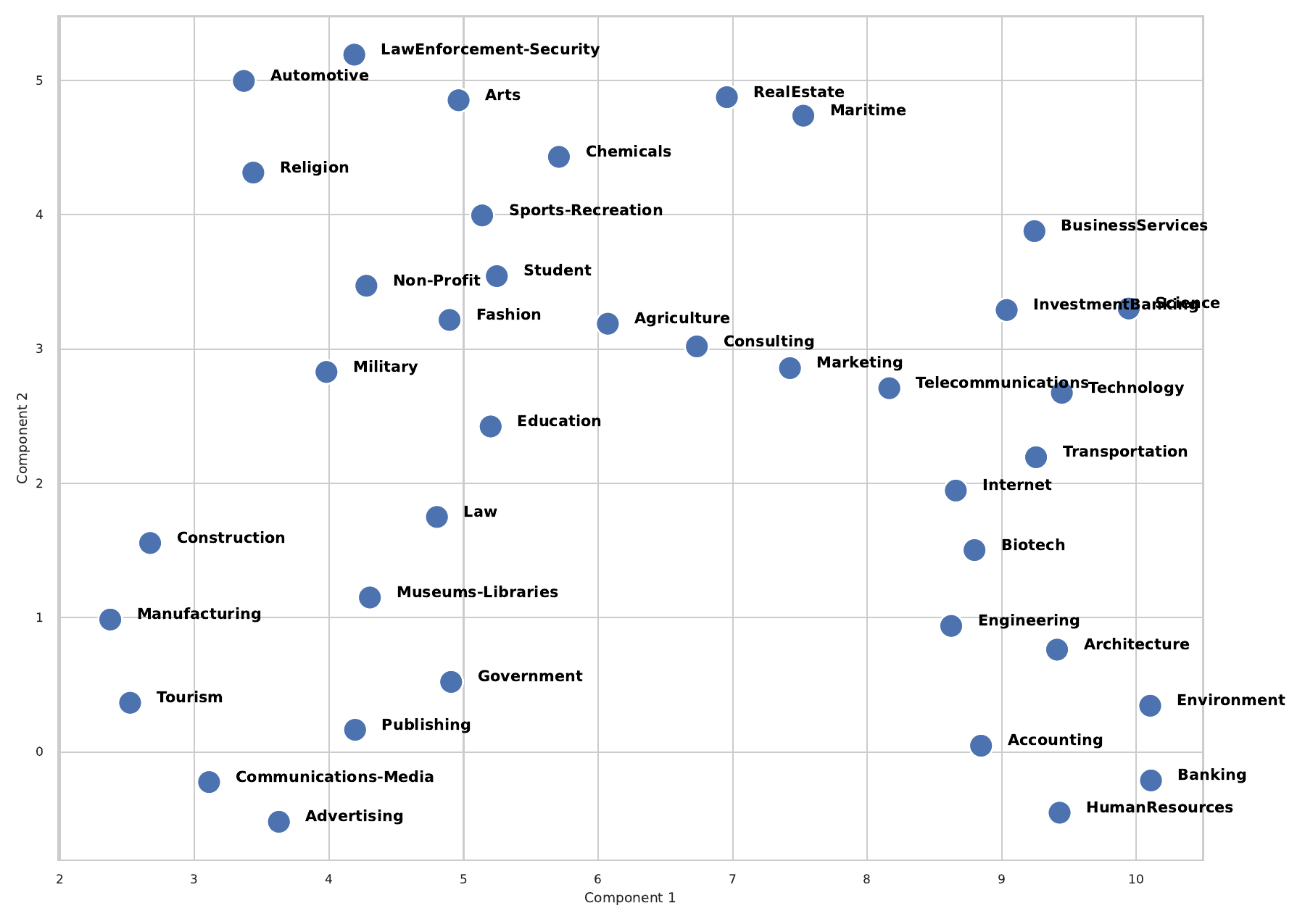}}
\caption{U-map projection of the combined embeddings for each occupation.}
\label{fig:umapoccupation}
\end{figure}

We also look at the occupation of the authors, for this purpose we present Fig.~\ref{fig:umapoccupation}, where the average embedding of all chunks with a given occupation is computed and projected. We want to find if authors with related occupations have similar writing styles encoded in their embeddings.

The bottom right corner seems more tech-related with subjects such as engineering, architecture, biotech and internet, mixed up with others such as environment or human resources. The bottom left corner has more communication-related occupations such as publishing, communications-media, advertising, tourism and, surprisingly, government. Finally the upper half is a mixed bag of occupations with no seeming connection whatsoever, such as chemicals, law enforcement or religion. Some are still related like sports-recreation, student and non-profit, but similarity seems almost arbitrary in that particular region. 

\section{Discussion}

We demonstrate state-of-the-art performance on authorship attribution and style understanding using public benchmark datasets from competitions. The model for the first case study, despite previous failure in zero-shot attribution, appears to form meaningful representations because mails by the same author lie close together in the embedding space. This is reinforced by the capacity of the model to create groups in the graph representation of the Gutenberg corpus or the sequential organization of age groups in the blog authorship dataset. Whatever the case study, there is overwhelming evidence that these encodings contain relevant information about the author.

However, the embeddings have side-effects with heavy impact on the model. The authorship embeddings, as stylistic numerical representations of authorship, also contain some information about the content. For example we have found that our embeddings are correlated with the type of writing, such as history books or fictional novels. The model is prone learn biases of the writers themselves, if most authors only write one type of book across their career, the model is going to assume that authors generally only write in one style, which is often the case for less prolific authors. It is also prone to capturing information about the author's age, gender and occupation of the author, as we have shown in the blog authorship dataset. However, what happens when an author switches styles, writes at an early age and at a later age or changes occupations, remains unexplored and would likely confound the model as the training data does not account for these dynamical scenarios.

Representation in the training set is extremely impactful. The correlation bias towards over-represented labels (fiction in the case of literature, or student for the blog occupation) points out that training representation is more important than previously anticipated, which is an issue considering that finding separate non-anonymous authors is difficult. Other worse represented labels (for example essays) have much worse embeddings in the resulting target space. A sample set of a thousand authors (Gutenberg corpus) can be limited, but this is also true for larger sample sizes (blog authorship). In the quantitative results, we find that training a model with a more diverse set of authors is beneficial to generalization, but still not enough. The general model succeeds in attribution for the three proposed datasets, and is able to produce meaningful representations of authorship in all of them.

\textit{To the best of our knowledge this work is the first to outline a method capable of zero-shot stylistic feature extraction for authorship}, which can be applied to attribution and style change tasks. While these results are very optimistic, we acknowledge that the training dataset is biased towards blogging language.

\section{Conclusions and future work} \label{sec:conclusions}
In this article we present a novel method for authorship representation called PART, the main contribution being the generated authorship embeddings and model. We formulate authorship embeddings as a numerical representation of an authors writing style and features, which is comparable with other authors and can be pooled with other works from the same writer to represent larger spans of text. 

We have observed the embeddings can detect authorship on unseen sets of authors and unseen domains. The model reaches competitive performance in tested PAN@CLEF tasks a zero-shot, outclasses baselines, achieving high accuracies in two of the analysed case studies, while dominating the proposed baselines on benchmark challenges. The utility of the authorship embeddings is measured qualitatively in three separate case studies. Every case study has access to different quality of information, and all are analyzed through with the same general-purpose authorship model. Texts written by the same author typically share regions in the embedding space generated by the model. The authorship embeddings include whether an author frequently writes poetry, is an engineer, has 17 years old or identifies as a female. In a related note, similar authors also share regions of the space, as shown with the Gutenberg Corpus.

Weaknesses of the model have appeared when experimenting with the model.
Despite having more than ten thousand authors from mixed origins the model is heavily biased toward some specific features we have detected. 
Adding training data from additional sources (social media, scientific journals, essays, etc.) could lead to a more robust generalization of the model. We also acknowledge the possibility of topic leakage into the model, as disentangling style and semantics completely is a hard task that we cannot fully prove to achieve.


\section*{Afterword}
We share our code at the following GitHub for replication: \url{https://github.com/jahuerta92/authorship-embeddings} 

\section*{Acknowledgments}

This work is part of the project PCI2022-134990-2 (MARTINI) of the CHISTERA IV Cofund 2021 program, funded by MCIN/AEI/10.13039/501100011033 and by the “European Union NextGenerationEU/PRTR”, supported by the Spanish Ministry of Science and Innovation under FightDIS (PID2020-117263GB-100) and MCIN/AEI/ 10.13039/501100011033/ and European Union NextGenerationEU/PRTR for XAI-Disinfodemics (PLEC2021-007681) grant, by European Comission under IBERIFIER - Iberian Digital Media Research and Fact-Checking Hub (2020-EU-IA-0252), and by "Convenio Plurianual with the Universidad Politécnica de Madrid in the actuation line of \textit{Programa de Excelencia para el Profesorado Universitario}". 
\bibliographystyle{unsrt}
\bibliography{references}

\end{document}